\documentclass[10pt,twocolumn,letterpaper]{article}

\usepackage{cvpr}
\usepackage{times}
\usepackage{epsfig}
\usepackage{graphicx}
\usepackage{amsmath}
\usepackage{amssymb}
\usepackage[pagebackref=true,breaklinks=true,letterpaper=true,colorlinks,bookmarks=false]{hyperref}
\hypersetup{
  pdftitle={A Controlled Study of Feature-Based Knowledge Distillation Across Student Designs},
  pdfauthor={Abhinand Balachandran and Praveen Prashant}
}

\cvprfinalcopy

\ifcvprfinal\fi
\begin{document}

\title{\textbf{A Controlled Study of Feature-Based\\
Knowledge Distillation Across Student Designs
}}

\author{Abhinand Balachandran\textsuperscript{*}\\
Georgia Institute of Technology\\
Atlanta, Georgia, USA\\
{\tt\small abalachandran7@gatech.edu}
\and
Praveen Prashant\textsuperscript{*}\\
Georgia Institute of Technology\\
Atlanta, Georgia, USA\\
{\tt\small pprashant3@gatech.edu}
}

\maketitle
\begingroup
\renewcommand\thefootnote{}
\footnotetext{\textsuperscript{*} Equal contribution. This work originated as the final project for Georgia Tech's CS 7643: Deep Learning course.}
\endgroup
\thispagestyle{empty}

\begin{abstract}
Knowledge distillation trains a smaller student to match the outputs of a larger teacher. Feature-based methods also align intermediate representations, but this extra constraint may affect students differently. We study this question on CIFAR-100 using a ResNet-50 teacher, a width-controlled CustomResNet family and MobileNetV2 as a cross-design comparison. For each student, we evaluate each feature method against a matched logit-KD run using the same teacher, optimizer settings, training schedule and seed. We repeat the main comparisons across multiple seeds.

Logit KD improved every tested student over its scratch baseline. Attention Transfer showed no clear relationship with size inside the CustomResNet family, but its average effect was negative for that family and positive for MobileNetV2. FitNets was below logit KD in all 15 paired runs. Within the constant-depth width sweep, its gap increased for wider students, although the different-depth $w{=}48$ student did not follow this trend. Finally, the same auxiliary coefficient produced different gradient scales across students, showing that a fixed coefficient does not create a uniform training condition.
\end{abstract}

\section{Introduction}
\subsection{Motivation}
Modern image classifiers are quite powerful, but they are also large. Architectures such as ResNet-50 contain about 24 million learned parameters. At inference, this translates to billions of arithmetic operations, even for classifying one image. Thus, for everyday use, it is important to make accurate image classifiers accessible to limited-compute devices such as personal mobile phones, health devices, and drones.

Knowledge distillation is one way around this. Instead of training a small network only on the ground-truth labels, we also train it to match the outputs of a large, already-trained ``teacher'' network~\cite{hinton2015distilling}. The intuition is that the teacher's softened class probabilities carry more information than a one-hot label, since they show how the teacher spreads its confidence across related classes (for example, that a leopard image looks somewhat like a tiger but nothing like a bicycle). This work studies how well that transfer works on CIFAR-100.

Specifically, we ask two questions. First, how much of the teacher's accuracy can a smaller model recover by matching the teacher's predictions? Second, does also matching patterns inside the two models improve accuracy, and does the answer change with the student's size or design? This matters in practice: our width-32 student is 21.6$\times$ smaller and requires 8.0$\times$ fewer multiply--accumulate operations than the teacher while finishing only 2.76 percentage points behind it. Knowing when an additional feature loss is useful can therefore avoid extra training complexity when deploying models on limited hardware.

\subsection{Current Practices and Limitations}
The original distillation method from Hinton et al.~\cite{hinton2015distilling} combines a cross-entropy term on the hard labels with a Kullback--Leibler divergence term that makes the student's softened class distribution match the teacher's. A temperature parameter~$T$ controls how much the output distributions are softened, and a weighting coefficient~$\alpha$ balances the two terms. We refer to this baseline throughout as \emph{logit distillation}.

Later methods try to transfer information from the teacher's internal representations as well. Attention Transfer~\cite{zagoruyko2017paying} computes a spatial attention map by summing squared activations across channels, then penalizes the $\ell_2$ distance between the teacher's and student's maps at matched stages. Since channels are summed, the networks do not need the same channel count. FitNets~\cite{romero2015fitnets} instead matches the full feature tensor at one intermediate layer. A learnable $1{\times}1$ convolution projects the student's features to the teacher's shape before their $\ell_2$ distance is minimized. We add either feature loss to the same logit-distillation objective with weight~$\beta$.

Prior work has shown that the teacher--student capacity gap matters. Cho and Hariharan~\cite{cho2019efficacy} found that a more accurate teacher can still hurt a small student, while Mirzadeh et al.~\cite{mirzadeh2020improved} proposed intermediate ``teacher assistant'' models to bridge large gaps. Broader studies have also compared feature- and logit-based methods across several architectures and capacity gaps~\cite{sarfraz2021knowledge,tian2020contrastive}. We focus on a more specific question: with the teacher, training recipe and logit-KD baseline held fixed, what extra effect does a feature objective have as the student changes? We test this across several widths in a CustomResNet family and use MobileNetV2 as a cross-design comparison. We also check the scale of~$\beta$, since the same numerical value can apply different training pressure when the loss reduction or gradient scale changes.

\subsection{Data}
We use CIFAR-100~\cite{krizhevsky2009learning} as our benchmark. Following the framework proposed by Gebru et al.~\cite{gebru2021datasheets}, we highlight the most relevant aspects of the dataset for our study.

\textbf{Composition.}
CIFAR-100 contains 60{,}000 colour images at 32$\times$32 pixels, divided into 100 mutually exclusive classes (everyday objects, animals, plants, vehicles, etc.) with exactly 600 images per class. The standard split assigns 50{,}000 images to training (500 per class) and 10{,}000 to testing (100 per class), giving a perfectly balanced distribution across classes.

\textbf{Creation.}
The dataset was curated by Krizhevsky in 2009 as a labeled subset of the 80 Million Tiny Images collection. The source images were gathered from the web, and class labels were assigned by paid student annotators.

\textbf{Preprocessing.}
We apply standard augmentation: random cropping (32$\times$32 with 4-pixel padding), random horizontal flips, and per-channel normalization. We do not create a separate validation split; every run uses the same fixed 200-epoch budget, and our headline results do not select the peak test epoch.

\textbf{Known limitations.}
CIFAR-100 was derived from the 80 Million Tiny Images collection, which its creators withdrew in 2020 after finding derogatory categories and offensive images~\cite{tinyimages2020withdrawal}. This history is relevant to the dataset's provenance even though we use only the standard CIFAR-100 split. Its 32$\times$32 resolution also means that our conclusions may not transfer directly to higher-resolution settings such as ImageNet. We chose CIFAR-100 because it is small enough to run a meaningful capacity sweep within our compute budget, yet difficult enough to leave a substantial gap between the teacher and smallest student. We downloaded it using \texttt{torchvision.datasets.CIFAR100}.

\section{Approach}

\subsection{Controlled Comparison and Models}
Our main contribution is a paired experimental setup rather than a new loss. For each student, we compare a feature objective with a matched logit-KD run while fixing the teacher, student initialization, optimizer settings, training schedule and seed. We repeat this over a ResNet width sweep and use MobileNetV2 as a cross-design check. Since it is our only non-ResNet student, this comparison can show an association with design but cannot prove the design caused the result.

We built the models in PyTorch using CIFAR-100 and components from \texttt{torchvision}~\cite{paszke2019pytorch,torchvision2016}. The ResNet-50 teacher uses a CIFAR stem with a 3$\times$3, stride-1 convolution and no max pooling. It reaches 78.54\% top-1. Label smoothing left its accuracy unchanged but made its $T{=}4$ distribution about three times flatter, so we distilled from the plain teacher.

Our CustomResNet uses residual BasicBlocks with stage widths $(w,2w,4w)$ and two spatial downsampling steps. Widths $w\in\{16,20,24,32\}$ all use $(3,3,3)$ blocks, which changes capacity while holding depth and structure fixed. The $w{=}48$ model uses $(2,2,2)$ blocks to provide a larger student at similar compute, so we mark it as off the controlled width curve in later analyses. We also test a MobileNetV2 student to see whether a different block design shows the same behaviour. These students span 0.278M to 2.352M parameters, compared with 23.705M for the teacher.

\subsection{Distillation Objectives}
For an input-label pair $(x,y)$, let $z_s$ and $z_t$ be the student and frozen teacher logits. All distilled students use
\begin{equation}
\begin{split}
\mathcal{L}_{\mathrm{KD}}={}&(1-\alpha)\,\mathrm{CE}(z_s,y)\\
&+\alpha T^2\,\mathrm{KL}(p_t^T\,\|\,p_s^T),
\end{split}
\label{eq:kd}
\end{equation}
where $p_m^T=\operatorname{softmax}(z_m/T)$. We fix $T{=}4$ and $\alpha{=}0.5$ after a small initial ablation. The $T^2$ factor keeps the soft-target gradient on a comparable scale as temperature changes.

Attention Transfer (AT) converts a feature tensor $f$ into a spatial map $A(f)=\sum_c f_c^2$, then normalizes the map to $Q(f)=A(f)/\lVert A(f)\rVert_2$. We match corresponding stages using
\begin{equation}
\mathcal{L}_{\mathrm{KD+AT}}=\mathcal{L}_{\mathrm{KD}}
+\beta\sum_j \operatorname{mean}\!\left[(Q(f_s^j)-Q(f_t^j))^2\right].
\label{eq:at}
\end{equation}
The three CustomResNet stages pair with the first three teacher stages at the same resolution. MobileNetV2's stem, middle and final features pair with the teacher stem, second stage and third stage. The maps and normalization have no learned parameters.

Our FitNets-style objective uses one intermediate hint. A learned $1{\times}1$ convolution $R$ maps student channels to teacher channels. Both tensors are $\ell_2$-normalized over channels at each spatial location, denoted by a bar, and
\begin{equation}
\begin{split}
\mathcal{L}_{\mathrm{hint}}={}&\operatorname{mean}_{n,h,w}
\left\lVert\overline{R(f_s)}-\overline{f_t}\right\rVert_2^2,\\
\mathcal{L}_{\mathrm{KD+Fit}}={}&\mathcal{L}_{\mathrm{KD}}+\beta\mathcal{L}_{\mathrm{hint}}.
\end{split}
\label{eq:fitnets}
\end{equation}
We pair the second ResNet stage or middle MobileNetV2 feature with the teacher's second stage. We train the student and regressor jointly for 200 epochs rather than using the original two-stage procedure. The student and regressor are learned; the teacher, temperature scaling and AT maps are fixed.

\subsection{Training Controls and Working Hypothesis}
Every student starts from a random initialization and uses the same 200-epoch recipe: batch size 128, SGD with learning rate 0.1, momentum 0.9, Nesterov acceleration, weight decay $5\times10^{-4}$ and cosine learning-rate decay. Within each pair, the initialization seed and training recipe are also the same. We use $\beta{=}1000$ for the main AT runs following the reference implementation. We also sweep $\{10,100,1000,10000\}$ on $w{=}16$. FitNets has a different raw loss scale, so its main value $\beta{=}3$ was chosen to place the hint term in a similar initial fraction of total loss as AT. It was not chosen by gradient parity.

A fixed numerical $\beta$ may not push every student by the same amount. We check this with the first-batch parity value
\begin{equation}
\beta^*=\frac{\lVert\nabla_{\theta_s}\mathcal{L}_{\mathrm{KD}}\rVert_2}
{\lVert\nabla_{\theta_s}\mathcal{L}_{\mathrm{aux}}\rVert_2},
\label{eq:parity}
\end{equation}
which equalizes the two student-gradient norms at initialization. We use it only to compare scales, not to claim that $\beta^*$ is optimal. Our initial grid suggested that AT might help students with spare capacity but overload smaller students. The width sweep and cross-design comparison test that idea.

\subsection{What Did Not Work First}
Several early implementations ran without errors but were not valid comparisons. First, stock MobileNetV2 downsampled a 32$\times$32 image to a 2$\times$2 final map because its strides were designed for ImageNet. Removing two later stride-2 operations restored an 8$\times$8 map and raised scratch accuracy from 68.72\% to 73.25\% without changing parameter count. We used this CIFAR-adapted version throughout.

Second, raw FitNets MSE let the regressor approach a zero-output shortcut. In a frozen-student probe, hint loss fell from 0.175 to 0.0073 even though the student could not learn. Channel normalization removed this shortcut. Our first AT reduction then summed rather than averaged over the 32$\times$32 map, making it 1024 times too large; normalized FitNets averaged over 512 channels, making it 512 times too small. These silent errors show that a $\beta$ value only makes sense together with the exact loss reduction.

\section{Experiments and Results}

\begin{figure*}[t]
\centering
\includegraphics[width=0.94\textwidth]{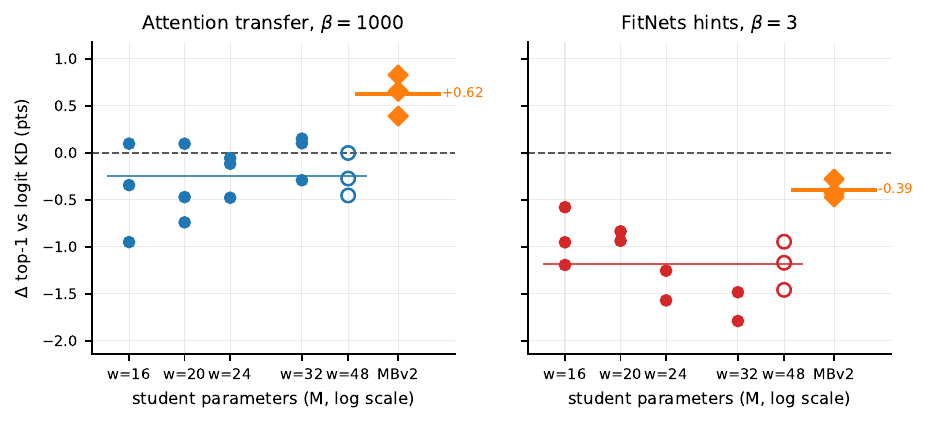}
\caption{Paired feature-loss effects for every seed. Points show the final-10-epoch difference from matched logit KD; horizontal segments show pooled means. Hollow markers denote the off-curve $w{=}48$. AT has three seeds per student. FitNets has three for $w{=}16$, $w{=}48$ and MobileNetV2 and two for the other widths.}
\label{fig:paired_deltas}
\end{figure*}

\begin{figure}[t]
\centering
\includegraphics[width=\columnwidth]{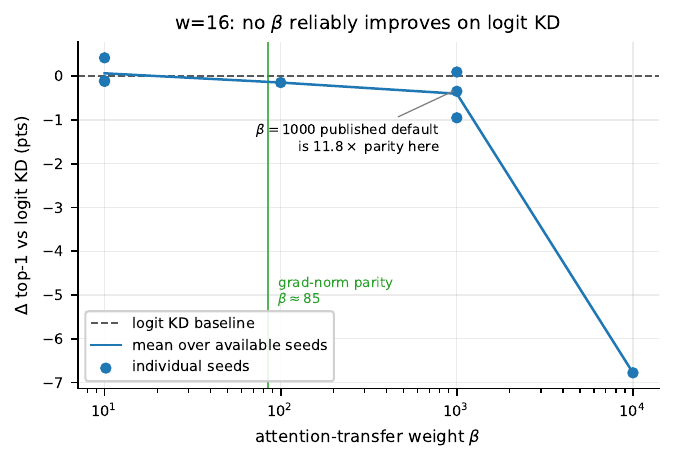}
\caption{AT coefficient sweep on $w{=}16$. Points are seeds and the line is their available mean. The $\beta{=}10$ and $1000$ arms have three seeds; $100$ and $10000$ have one. Low weight is neutral after replication, while high weight fails clearly.}
\label{fig:beta_sweep}
\end{figure}

\begin{figure*}[t]
\centering
\includegraphics[width=0.76\textwidth]{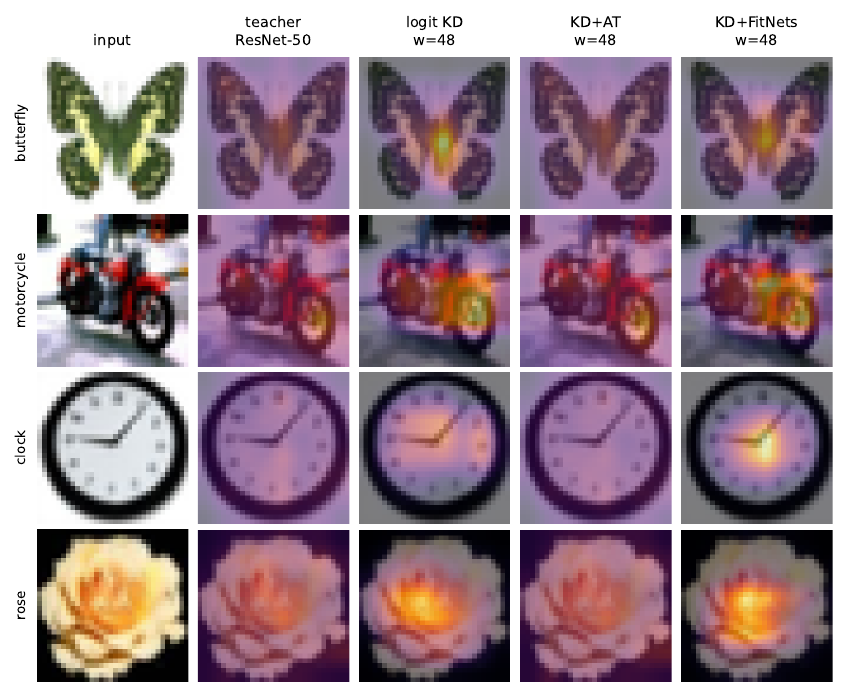}
\caption{Normalized spatial attention at the deepest matched stage. Every model predicts each row correctly, using a shared scale per row. Examples come from four predeclared groups among 2,025 jointly correct images and require at least 85\% lower teacher-map error under AT. AT was closer than logit KD for all 2,025 images, so the rows illustrate an aggregate result.}
\label{fig:attention_maps}
\end{figure*}

\subsection{Reporting Protocol}
We measure success by test top-1 accuracy, summarized as the mean and sample standard deviation over the final 10 epochs. The standard deviation describes fluctuation within one run, not uncertainty across seeds. We avoid peak test accuracy because it uses the test set for model selection and raises scores by roughly 0.1 to 0.5 points.

We define $\Delta_{\mathrm{KD}}=\mathrm{KD}-\mathrm{scratch}$ and $\Delta_{\mathrm{aux}}=(\mathrm{KD+aux})-\mathrm{KD}$. Each auxiliary comparison uses the same seed and recipe as its KD baseline. We state $n$ because some secondary settings have fewer repeats. Run variation is about 0.3 to 0.5 points, so a difference between two single runs can carry roughly 0.6 to 0.9 points of noise. We treat single-seed effects below one point as observations to test, not conclusions.

\subsection{Main Grid}
The seed-42 grid in Table~\ref{tab:main_grid} covers scratch, logit KD, AT and FitNets. Initial FitNets runs used parity-based $\beta$ values of 8, 100 and 6 for $w{=}16$, MobileNetV2 and $w{=}48$, producing very different loss shares. We reran them with the final loss and a common $\beta{=}3$, which gave the most balanced initial shares. The table reports these replacement runs.

The teacher reached $78.54\pm0.08$ top-1. Logit KD improved all three students over scratch by $+0.46$, $+2.22$ and $+1.09$ points for $w{=}16$, MobileNetV2 and $w{=}48$. Their teacher gaps were 8.71, 5.29 and 3.01 points, so the largest KD gain occurred in the middle rather than tracking capacity.

\begin{table}[h!]
\centering
\small
\setlength{\tabcolsep}{2.2pt}
\begin{tabular}{lccc}
\hline
Method & $w{=}16$ & MobileNetV2 & $w{=}48$ \\
\hline
Scratch & $69.84{\pm}0.12$ & $73.25{\pm}0.15$ & $75.54{\pm}0.11$ \\
Logit KD & \textbf{$70.29{\pm}0.08$} & $75.47{\pm}0.10$ & \textbf{$76.63{\pm}0.12$} \\
KD+AT & $69.95{\pm}0.11$ & \textbf{$76.12{\pm}0.12$} & \textbf{$76.63{\pm}0.10$} \\
KD+FitNets & $69.34{\pm}0.19$ & $75.19{\pm}0.09$ & $75.17{\pm}0.18$ \\
\hline
\end{tabular}
\caption{Main comparison at seed 42. The $w{=}16$ and $w{=}48$ columns are CustomResNets. AT uses $\beta{=}1000$ and FitNets uses $\beta{=}3$. Values are the mean $\pm$ sample standard deviation of test top-1 accuracy (\%) over the final 10 epochs.}
\label{tab:main_grid}
\end{table}

Relative to KD, AT changed $w{=}16$, MobileNetV2 and $w{=}48$ by $-0.34$, $+0.65$ and $0.00$ points. This suggested that smaller students might struggle with the extra constraint. FitNets was below KD by $-0.95$, $-0.28$ and $-1.46$ points. These were seed-42 observations, and the MobileNetV2 AT gain was below our single-seed noise threshold. They formed hypotheses for the width, seed and loss-scale checks below.

\subsection{Testing the Capacity Hypothesis}
We first filled the gap between $w{=}16$ and $w{=}48$ with constant-depth students. At seed 42, the AT deltas for $w{=}16,20,24,32$ were $-0.34$, $-0.74$, $-0.48$ and $+0.15$ points. This was not the smooth improvement with capacity suggested by the main grid. We then repeated every AT width at three seeds. Across the 15 paired CustomResNet runs, the relationship between AT effect and log parameter count was weak and not statistically clear ($r{=}0.31$, $p{=}0.27$). We therefore found no evidence that larger students benefited more from AT.

\subsection{A Design-Associated Sign Reversal}
Although the width ordering disappeared, the multi-seed runs revealed a clearer split in Fig.~\ref{fig:paired_deltas}(a). AT reduced accuracy on the CustomResNet family by $0.25$ points on average, with 11 of 15 paired runs below zero. The 95\% confidence interval was $[-0.43,-0.07]$ ($p{=}0.011$). MobileNetV2 instead gained $0.62\pm0.22$ points ($n{=}3$, $p{=}0.039$), with all three seeds positive.

For a direct comparison, we averaged the five ResNet widths within each seed and compared that mean with MobileNetV2 at the same seed. The difference was $+0.87\pm0.30$ points ($t{=}5.04$, $p{=}0.037$). This tests the difference between the groups directly rather than comparing two separate significance results. We call it a \emph{student-design-associated} sign reversal. It does not prove that architecture caused the reversal because feature pairing and effective loss strength also changed. As a partial check, $w{=}48$ was closest to MobileNetV2 in gradient scale at $6.8\times$ versus $5.2\times$ parity, yet their AT effects were still on opposite sides of zero ($-0.24$ versus $+0.62$; paired contrast $p{=}0.021$). This makes loss scale a less complete explanation, but it does not rule it out.

\begin{table}[h!]
\centering
\small
\setlength{\tabcolsep}{3.2pt}
\begin{tabular}{lccc}
\hline
Comparison & $n$ & Mean $\Delta$ & $p$ \\
\hline
AT, ResNet family & 15 & $-0.25{\pm}0.33$ & 0.011 \\
AT, MobileNetV2 & 3 & $+0.62{\pm}0.22$ & 0.039 \\
AT, design contrast & 3 & $+0.87{\pm}0.30$ & 0.037 \\
FitNets, ResNet family & 12 & $-1.18{\pm}0.35$ & $<0.0001$ \\
FitNets, MobileNetV2 & 3 & $-0.39{\pm}0.10$ & 0.021 \\
\hline
\end{tabular}
\caption{Multi-seed paired effects in accuracy points, shown as mean $\pm$ sample standard deviation. For the design contrast, $n$ denotes seed blocks; otherwise it denotes paired runs.}
\label{tab:paired_summary}
\end{table}

\subsection{The Auxiliary Weight Matters}
The $w{=}16$ AT sweep shows why a coefficient from another setup may not work directly in ours. At seed 42, $\beta{=}10,100,1000,10000$ gave deltas of $+0.42$, $-0.14$, $-0.34$ and $-6.78$ points. The apparent gain at $\beta{=}10$ did not hold across seeds: its three-seed mean was $+0.07\pm0.31$ ($t{=}0.38$). Its paired advantage over $\beta{=}1000$ was also unclear at three seeds ($+0.47\pm0.58$, $t{=}1.40$). Lowering $\beta$ made AT roughly neutral on this student, not reliably helpful.

As Fig.~\ref{fig:beta_sweep} shows, gradient-norm parity was near $\beta^*{=}85$ for $w{=}16$, making the default $\beta{=}1000$ about $11.8\times$ parity. The same default was $6.8\times$ parity for $w{=}48$ and $5.2\times$ for MobileNetV2. Thus, the same $\beta$ did not create the same training pressure. This check explains why coefficient scale must be reported, but it does not tell us what caused the cross-design reversal.

\subsection{FitNets Gives the Strongest Negative Result}
FitNets was below its paired KD baseline in all 12 CustomResNet runs. Its average change was $-1.18\pm0.35$ points ($t{=}-11.70$, $p{<}0.0001$). MobileNetV2 also lost $0.39\pm0.10$ points ($n{=}3$, $p{=}0.021$), making all 15 runs negative across the two student designs. This differs sharply from AT, whose sign changed for MobileNetV2. The contrast is visible across the two panels of Fig.~\ref{fig:paired_deltas} and in Table~\ref{tab:paired_summary}.

Within the controlled $(3,3,3)$ width sweep from $w{=}16$ through $w{=}32$, the FitNets gap grew with capacity ($r{=}-0.81$, $p{=}0.008$, $n{=}9$). This is the reverse of our original capacity argument. The $w{=}48$ student did not follow the trend: its mean gap was $-1.19$ rather than falling below $w{=}32$ at $-1.64$. Since $w{=}48$ also changes depth, we left it out of the controlled correlation. Including it weakens the relationship to $r{=}-0.53$ ($p{=}0.079$).

\subsection{Loss Dynamics and Qualitative Evidence}
Both feature losses became worse as their coefficient increased on $w{=}16$. FitNets changed accuracy by $-0.05$, $-0.95$ and $-2.68$ points at $\beta{=}0.3,3,30$. AT followed the same broad response from neutral at low weight to severe damage at high weight. Their loss shares also drifted during training. AT began near 18\% for every ResNet width but ended between 24.6\% and 48.4\%; FitNets showed a similar fan. The share rose because the KD term shrank faster than the feature residual. It does not by itself explain accuracy because students with similar final shares had different outcomes.

Figure~\ref{fig:attention_maps} gives the qualitative view. All four models classify every displayed image correctly. AT closely matches the teacher's deepest paired spatial map, reducing the selected-row mean map error from $7.260\times10^{-3}$ for logit KD to $2.283\times10^{-4}$. Yet its replicated $w{=}48$ accuracy is still $0.24$ points below KD. FitNets has map error $8.310\times10^{-3}$ and loses $1.19$ points, although it directly optimizes a different feature at an earlier layer. The main result is that AT can match the intended representation much more closely without improving classification. Overall, our original capacity hypothesis failed. What we can support instead is a student-design-associated change in AT's sign, while FitNets is consistently worse than logit KD in our tested setting.

\section{Discussion and Limitations}
Our experiments show that adding an auxiliary feature loss on top of logit distillation does not give one consistent result. FitNets hurt every student we tested, while AT hurt the CustomResNet family on average but helped MobileNetV2. We originally hypothesized that student capacity would explain these differences. However, our width sweep and seed replications do not support a capacity-based trend.

Our initial single-seed results suggested that AT hurts small students and helps larger ones, consistent with the intuition that small networks may struggle with an additional constraint. This pattern did not hold consistently across more widths and seeds. Across the 15 paired CustomResNet runs, the relationship between AT's effect and model size was weak ($r{=}0.31$, $p{=}0.27$). A clearer difference appeared between student designs. AT reduced CustomResNet accuracy by 0.25 points on average but improved MobileNetV2 by 0.62 points, and the direct seed-blocked difference was 0.87 points ($p{=}0.037$). We describe this result as design-associated rather than architecture-caused. MobileNetV2 is our only non-ResNet student, and its feature pairing and effective loss strength also differ.

One possible explanation was that holding $\beta$ fixed places different levels of constraint on each student. The default $\beta{=}1000$ is $11.8\times$ the parity scale for $w{=}16$ but only $5.2\times$ for MobileNetV2. However, our results do not show that this caused the sign reversal. Lowering $\beta$ from 1000 to 10 made AT close to neutral on $w{=}16$, not reliably positive, and the difference between these settings was not clear with three seeds. Also, $w{=}48$ and MobileNetV2 have similar parity multiples but AT still had opposite signs for them. Gradient parity is therefore useful for checking scale, but it does not give us the best $\beta$ or explain the full result.

For FitNets, we observe a consistent deficit in all 15 paired runs. Within the constant-depth sweep from $w{=}16$ to $w{=}32$, this deficit grew with student width ($r{=}-0.81$, $p{=}0.008$). The different-depth $w{=}48$ student did not follow the trend, and including it weakens the relationship to $r{=}-0.53$ ($p{=}0.079$). One reason for the constant-depth result may be that a wider $1{\times}1$ regressor constrains a higher-rank subspace, forcing wider students to reproduce more of the teacher's representation. We have not verified this and present it only as a possible explanation. The FitNets loss share also rises over this width sweep, so our experiments cannot separate the two explanations.

The attention maps show why a lower feature loss should not be treated as better classification. AT reduced the teacher-map error by about $32\times$ on the selected rows, yet its replicated $w{=}48$ accuracy was still 0.24 points below KD. The loss shares also changed during training because the KD term fell faster than the feature residual. Since similar final shares led to different accuracy changes, we do not claim that this drift caused the results.

\textbf{Limitations.} We used one dataset, one teacher and one 200-epoch schedule without a validation split. Train accuracy ranged from 83.5\% to 99.9\%, so the students also had different generalization gaps. MobileNetV2 has only three seeds, while several secondary FitNets and $\beta$ settings have one or two. We therefore treat the statistical tests as exploratory. Future work should test more student designs at similar capacity and better-matched loss strength, then repeat the study at higher resolution.

\section{Conclusion and Future Work}

We asked whether feature matching improves accuracy beyond logit KD and whether its benefit depends on student size. Logit KD improved every main-grid student, but our capacity hypothesis for the feature losses did not hold. AT was negative on average for the CustomResNet family and positive for MobileNetV2. This gives us a design-associated result, but our experiments cannot show that architecture caused it. FitNets was below logit KD in all 15 paired runs.

The main lesson is that a fixed $\beta$ does not create the same training condition for every student. Small choices in loss reduction can also change its scale by orders of magnitude. In our experiments, successfully matching the intended representation was not enough for a feature loss to improve accuracy. Future work should test more student designs at similar capacity and better-matched loss strength before making an architectural claim.

{\small
\bibliographystyle{ieee_fullname}
\bibliography{references}
}

\end{document}